\documentclass[conference]{IEEEtran}
\usepackage{blindtext, graphicx}
\usepackage{listings}
\usepackage[numbers]{natbib}

\usepackage[colorlinks=true]{hyperref}

\usepackage{array}
\usepackage{multirow}
\usepackage{siunitx}
\usepackage{color,soul}
\usepackage{caption}
\usepackage{subcaption}

\usepackage{booktabs}
\usepackage{hyperref}
\usepackage[utf8]{inputenc}
\usepackage{mathtools}
\usepackage{adjustbox}
\usepackage{amsmath}
\usepackage{amssymb}
\usepackage[mathlines]{lineno}
\usepackage{lineno}

\usepackage{algorithm} 
\usepackage{algpseudocode}

\usepackage[figurename=Fig.]{caption}

\usepackage{array}
\usepackage{multirow}
\usepackage[symbol]{footmisc}

\title{Enhanced Droplet Analysis Using Generative Adversarial Networks}

\author{\IEEEauthorblockN{Tan-Hanh Pham}
\IEEEauthorblockA{Department of Mechanical and\\
Civil Engineering, Florida Institute of\\
Technology, USA\\
Email: tpham2023@my.fit.edu}
\and

\IEEEauthorblockN{Kim-Doang Nguyen\textsuperscript{*}}
\IEEEauthorblockA{Department of Mechanical and\\
Civil Engineering, Florida Institute of\\
Technology, USA\\
Email: knguyen@fit.edu}}
\begin{document}

\maketitle
\thispagestyle{empty}
\pagestyle{empty}

\begin{abstract}
Precision devices play an important role in enhancing production quality and productivity in agricultural systems. Therefore, the optimization of these devices is essential in precision agriculture. Recently, with the advancements of deep learning, there have been several studies aiming to harness its capabilities for improving spray system performance. However, the effectiveness of these methods heavily depends on the size of the training dataset, which is expensive and time-consuming to collect. To address the challenge of insufficient training samples, we developed an image generator named DropletGAN to generate images of droplets. The DropletGAN model is trained by using a small dataset captured by a high-speed camera and capable of generating images with progressively increasing resolution. The results demonstrate that the model can generate high-quality images with the size of $1024\times1024$. The generated images from the DropletGAN are evaluated using the Fréchet inception distance (FID) with an FID score of 11.29. Furthermore, this research leverages recent advancements in computer vision and deep learning to develop a light droplet detector using the synthetic dataset. As a result, the detection model achieves a 16.06\% increase in mean average precision (mAP) when utilizing the synthetic dataset. To the best of our knowledge, this work stands as the first to employ a generative model for augmenting droplet detection. Its significance lies not only in optimizing nozzle design for constructing efficient spray systems but also in addressing the common challenge of insufficient data in various precision agriculture tasks. This work offers a critical contribution to conserving resources while striving for optimal and sustainable agricultural practices.


\end{abstract}


\section{Introduction}
\label{sec.intro}

\subsection{Motivation}

\begin{figure*}[h]
\begin{center}
\includegraphics[width= 1\textwidth]{ 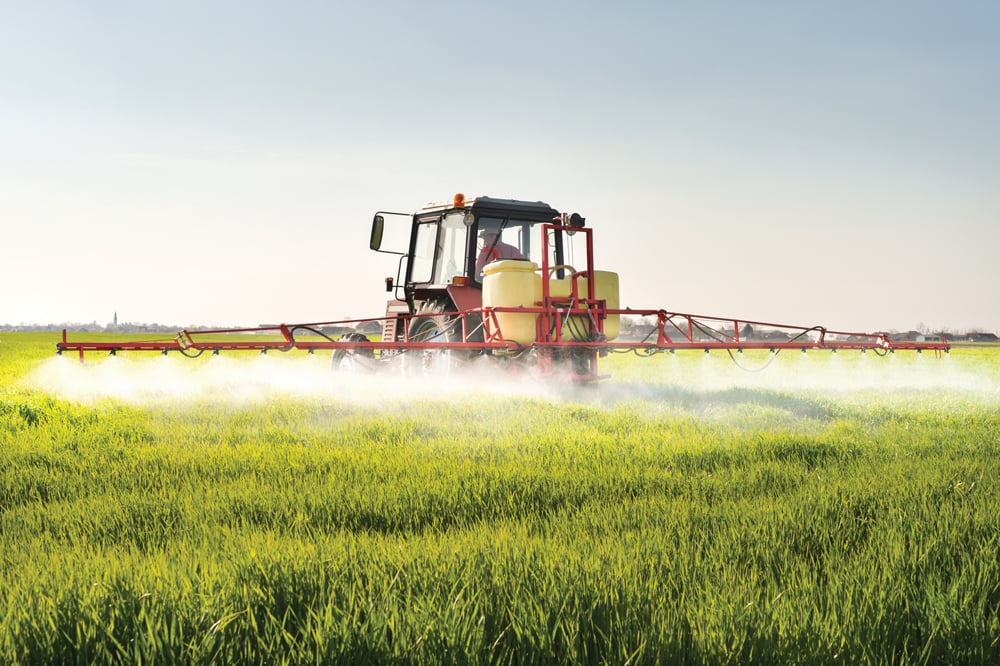}
\end{center}
\caption{Application of spray systems in agriculture (image is taken from www.grainews.ca)}
\label{fig.spray} 
\end{figure*}

In the current era of global environmental transformations caused by climate change, it is imperative to prioritize global sustainability and food security to ensure the well-being and support the livelihoods of humanity. In agriculture, numerous innovations leverage modern technologies to enhance productivity while minimizing environmental impact. Beyond the natural conditions of the environment, technology emerges as a pivotal player in agriculture, empowering the system to conserve resources and operate with greater efficiency. One noteworthy innovation is the use of spray systems in nurturing plants, considered a highly effective priority tool. These systems, when integrated with other technologies, not only save farmers time but also enhance synchronization on their farms. As depicted in Figure \ref{fig.spray}, a spray system can uniformly distribute substances into plants, facilitating optimal nutrient absorption or distributing pesticides to help crops minimize diseases.

Precision spraying systems play a crucial role in modern agriculture due to their impact on droplet drift and efficacy \citep{mahmud2021opportunities, seol2022field}. There are several kinds of sprayers such as air blast sprayers \citep{peterson1994tunnel}, steam-powered sprayers \citep{brann1956apparatus}, smart sprayers \citep{chen2010development}, etc. The air blast sprayer, for instance, comes in various configurations, including axial-fan, cannon, tunnel, tower, and custom designs. While axial-fan sprayers, with their large radial spray volume, have been the mainstay for tree fruit orchards, they have limitations in modern, intensive settings. With axial-fan sprayers, they are susceptible to significant off-target drift and chemical waste due to their broad spray pattern \citep{mahmud2021opportunities}.

Smart sprayers, on the other hand, are a new generation of sprayers that leverage advanced technologies like machine learning and computer vision to improve the efficiency and precision of chemical applications in agriculture \citep{qu2024deep}. Unlike traditional sprayers that apply a uniform amount of chemicals across a field, smart sprayers can adjust the application rate based on real-time data. This data can include information about the presence and type of weeds, the size and health of crops, and even weather conditions \citep{chen2022characteristics}. By targeting only the areas that need treatment, smart sprayers can significantly reduce waste and environmental impact, while also minimizing costs for farmers. Additionally, some smart sprayers can detect and target individual plants, further reducing chemical use and exposure for workers \citep{altalak2022smart}.

Optimizing spray nozzles is crucial for enhancing precision in spraying systems. This optimization involves adjusting nozzle parameters to change the characteristics of released droplets and make them suitable to each crop's unique needs and environmental conditions \citep{ahmad2021advancements}. There are several methods to measure the droplet characteristics. Traditional optical measurement methods include high-speed photography and laser imaging \citep{balewski2010experimental, kawaguchi2002size}. For example, one common method estimates droplet size by spraying colored water on a white sheet and then analyzing the resultant patterns \citep{schmandt2011diffuser, goddeeris2006light}. These methods are often highly manual, require delicate setups, and are associated with high operating costs.

Recently, with the advent of advanced machine learning, \cite{acharya2022ai} developed a novel tool to estimate the size and velocity of droplets sprayed from nozzles. However, the methods are not suitable for measuring the droplet deposition on stems and leaves in field spraying. Moreover, \cite{wang2021smart} developed a lightweight droplet detector that can be easily incorporated into sensing systems. \cite{yang2022droplet} proposed a droplet deposition characteristics detection method based on deep learning. Despite these advancements, these models are heavily dependent on the size of the datasets used for model training. However, gathering data to train machine learning models in most precision agriculture applications remains a time-consuming, labor-intensive, and expensive process \citep{bhat2021big, malek2022review, qu2024deep}.

Motivated by these challenges, this paper utilizes deep-learning techniques to introduce an alternative method that facilitates the generation of a substantial droplet dataset. This method is built upon the foundation of a generative adversarial neural network. Additionally, to assess the effectiveness of the proposed method, we establish a deep learning-based droplet detector by leveraging recent advances in computer vision.

\subsection{Literature review}

Over the past decade, deep learning has been developing rapidly with appealing applications across various domains, including finance, medicine, education, and agriculture \citep{ozbayoglu2020deep, pham2024soil, pham2023seunet, bhardwaj2021application, pham2023deep}. In agriculture, deep learning techniques have played a crucial role in addressing complex challenges such as crop detection, fruit segmentation, and produce classification \citep{hameed2018detection, shanmugam2020automated, liu2023automated}. For instance, \cite{tian2019apple} applied deep-learning algorithms for the detection and segmentation of apples. For crop detection, \cite{hamuda2018improved} developed an algorithm capable of tracking crops in sunny conditions. For weed detection, \cite{czymmek2019vision} proposed a real-time detection method using deep learning. For harvesting, \cite{divyanth2022detection} proposed a deep learning-based object detection for detecting coconut clusters.

Beyond the application of machine learning in crop treatment, it also plays an important role in obtaining measurements for optimizing the design of agricultural devices. For example, work in \cite{acharya2022ai, acharya2023deep} employed deep-learning techniques to estimate the spray cone angles and the characteristics of the droplets. Moreover, \cite{choi2022deep} proposed a deep-learning approach for discerning the droplet jetting dynamics in single-jet printing processes. However, a major challenge in utilizing these novel methods lies in data collection. The process often requires complex and expensive setups, incurring significant labor costs.  As a result, the accuracy and reliability of the resulting machine learning models heavily depend on the quality and quantity of data within the training dataset.

To resolve this data-size bottleneck and hence increase the performance of deep learning models, there are several techniques we can use, for example collecting more data, transfer learning, data augmentation, and synthetic data generation \citep{shorten2019survey}. For data augmentation, we can apply transformations to existing data to create new samples, including rotations, flips, zooms, or changes in brightness and contrast \citep{shorten2019survey}. Generally, the performance of the model could be improved by applying the data augmentation techniques. Compared to the data augmentation techniques, synthetic data generation can create new samples based on the original samples. This can be especially beneficial when creating novel patterns or dealing with data from class imbalance. 

Generative Adversarial Networks (GANs) have revolutionized content and model generation by introducing a novel framework for training generative models \citep{goodfellow2014generative}. GANs are usually constructed by two models: a generator and a discriminator. The generator produces images from noise to fool the discriminator, while the discriminator aims to distinguish real images from generated ones. This mechanism is trained until the images generated by the generator resemble real images.

Despite the success of the early GAN models, GAN training can be challenging due to model collapse and instability. Therefore, subsequent GAN models have been developed \citep{radford2015unsupervised, salimans2016improved, liu2016coupled, zhu2017unpaired} to address these issues. Although the performance of the models has been improved, the quality of generated images remains limited. Towards the end of 2017, researchers from NVIDIA AI released Progressive GAN (ProGAN) \citep{karras2017progressive}, which laid the foundation for later works such as StyleGAN \citep{karras2019style} and MSG-GAN \citep{karnewar2020msg}.

In addition to GANs, Variational Autoencoder (VAEs) and Flow-based Deep Generative Models are commonly used for image generation \citep{hinton2006reducing, bond2021deep}. For example, in VAEs, the models compress an input into a latent space representation, capturing its essence, and then attempt to reconstruct the original data from this compressed version. In VAEs, the twist lies in how the models represent the data in latent space as a probability distribution. This probabilistic approach allows VAEs to not only reconstruct data but also to sample new points from the learned distribution, effectively generating novel data that resembles the training data. Despite its efficiencies and performances, VAEs can sometimes struggle with capturing complex data structures. 

In 2020, the field of diffusion models saw a significant advancement with the proposal of denoising diffusion probabilistic models (DDPM) by \cite{ho2020denoising}. Prior to this, several generative models based on the core concept of diffusion had been introduced, such as those by \citep{sohl2015deep, song2019generative}. In diffusion models, we start by gradually adding noise to a real image from the training data. This process essentially corrupts the image with increasing levels of noise across a series of steps. Then, we train a diffusion model to learn the inverse process, which is how to remove the added noise and recover the original clean image gradually. Once trained, these models can generate entirely new images by starting with pure noise and progressively removing noise following the learned denoising process. With each step, the noise is reduced, revealing more and more image details. Diffusion models distinguish themselves from GANs and VAEs by utilizing a fixed training procedure and employing high-dimensional latent variables. 

Diffusion models have emerged as a powerful tool for generating different kinds of data. For image generation, DALL-E 2, developed by OpenAI, is one the popular models that create high-fidelity images from text descriptions \citep{ramesh2022hierarchical}. Similar to DALL-E 2, Imagen is another powerful text-to-image diffusion model developed by Google AI \citep{saharia2022photorealistic}. In addition, Stable Diffusion is also built using denoising diffusion probabilistic techniques \citep{rombach2021highresolution}. Beyond image generation, active research is exploring diffusion models' capabilities in diverse areas such as generating videos, training time improvement, speech generation, and image reconstruction \citep{xing2023survey, po2023state, hoover2023memory, liu2023generative}.

The application of generative models in agriculture has become an alternative to traditional methods, leading to a substantial enhancement in precision agriculture \citep{ward2018deep, espejo2021combining, toda2020training}. For example, \cite{valerio2017arigan} created a new dataset of artificial plants for plant phenotyping by using a generative model. \cite{toda2020training} demonstrates that they can achieve an accuracy of 95\% using only synthetic data for crop seed phenotyping. \cite{chen2023deep} proposed an augmentation technique using diffusion models for weed recognition. 

Despite its performance, diffusion-based models require a large dataset, and it takes a long time to train, especially with complex structured datasets. While VAEs and flow-based models offer alternative approaches, they can struggle to capture complex details and generate high-resolution images. Therefore, we proposed a droplet generator using GAN architectures, named DropletGAN. Our approach aims to save time and resources compared to traditional data collection methods for droplets.

\subsection{Contributions}
In this study, to address the bottleneck of data limitation and droplet analysis, we propose DropletGAN, a method used to generate synthetic droplet images. Since droplets are small, generating small objects that have a similar appearance as droplets poses a tremendous challenge for a generative model. Therefore, our approach involves generating high-resolution images from low-resolution ones by leveraging the progressive generative methodology. Additionally, this paper establishes a novel droplet detection method using a cutting-edge deep-learning algorithm trained on AI-generated synthetic data to measure their size.

The key contributions of our work include:
\begin{itemize}
    \item We developed DropletGAN, which is capable of generating high-quality droplet images.
    \item The Fréchet inception distance (FID) score of the generated images is 11.29.
    \item This work introduces a real-time droplet detection model, trained on synthetic data, capable of identifying small droplets.
    \item We demonstrate that the integration of synthetic droplet images improves the detector performance by 16.06\%.
    \item This work lays the foundation for integrating modern generative AI methods to address other tasks in agriculture.
\end{itemize}

The rest of the paper is organized as follows: Section \ref{sec.data} describes the dataset and data processing. Section \ref{sec.method} elaborates on the methodologies of deep learning algorithms behind the droplet image generator. In this section, we also discuss the droplet detection algorithm. The model training and metrics for evaluating the performance of the models are explained in Section \ref{sec.metrics}. Section \ref{sec.result} discusses the results and compares the effects of using a synthetic dataset. Finally, Section \ref{conclusion} provides concluding remarks on this work.

\section{Experimental Dataset}
\label{sec.data}
\begin{figure}[h]
\begin{center}
\includegraphics[width= 0.485\textwidth]{ 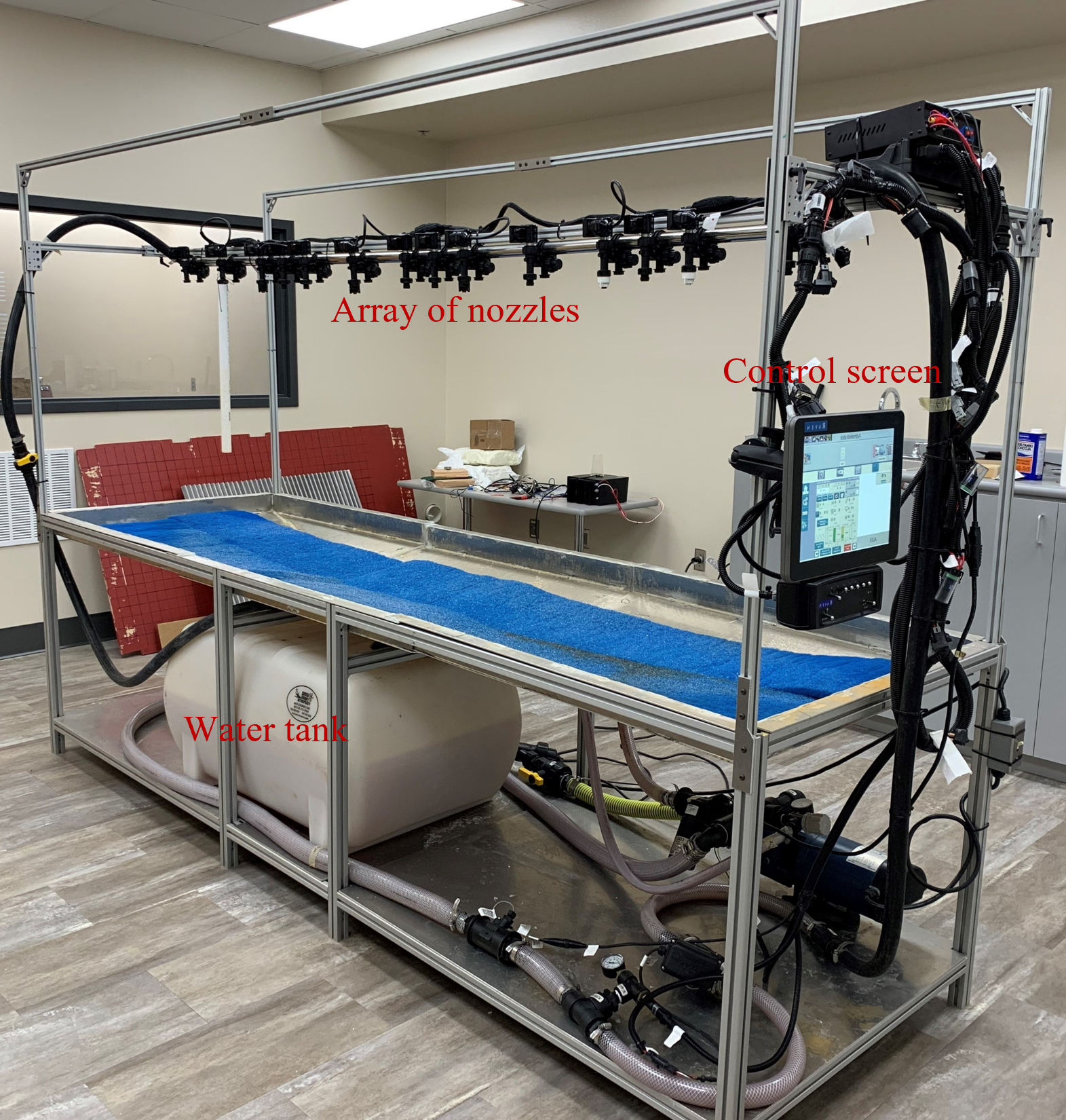}
\end{center}
\caption{Experimental setup of a spray system for droplet generation.}
\label{fig.experiment_setup} 
\end{figure}

Inspired by our prior work \citep{acharya2022ai}, we gathered data through an experimental setup with an array of crop spray nozzles capable of producing various spray patterns Fig. \ref{fig.experiment_setup}. In the experiments, we captured images of droplets with a resolution of $640 \times 640$ sprayed by the nozzles using a high-speed camera operating at 2000 frames per second to construct a droplet database. 

\begin{figure*}[h]
\begin{center}
\includegraphics[width= 0.8\textwidth]{ 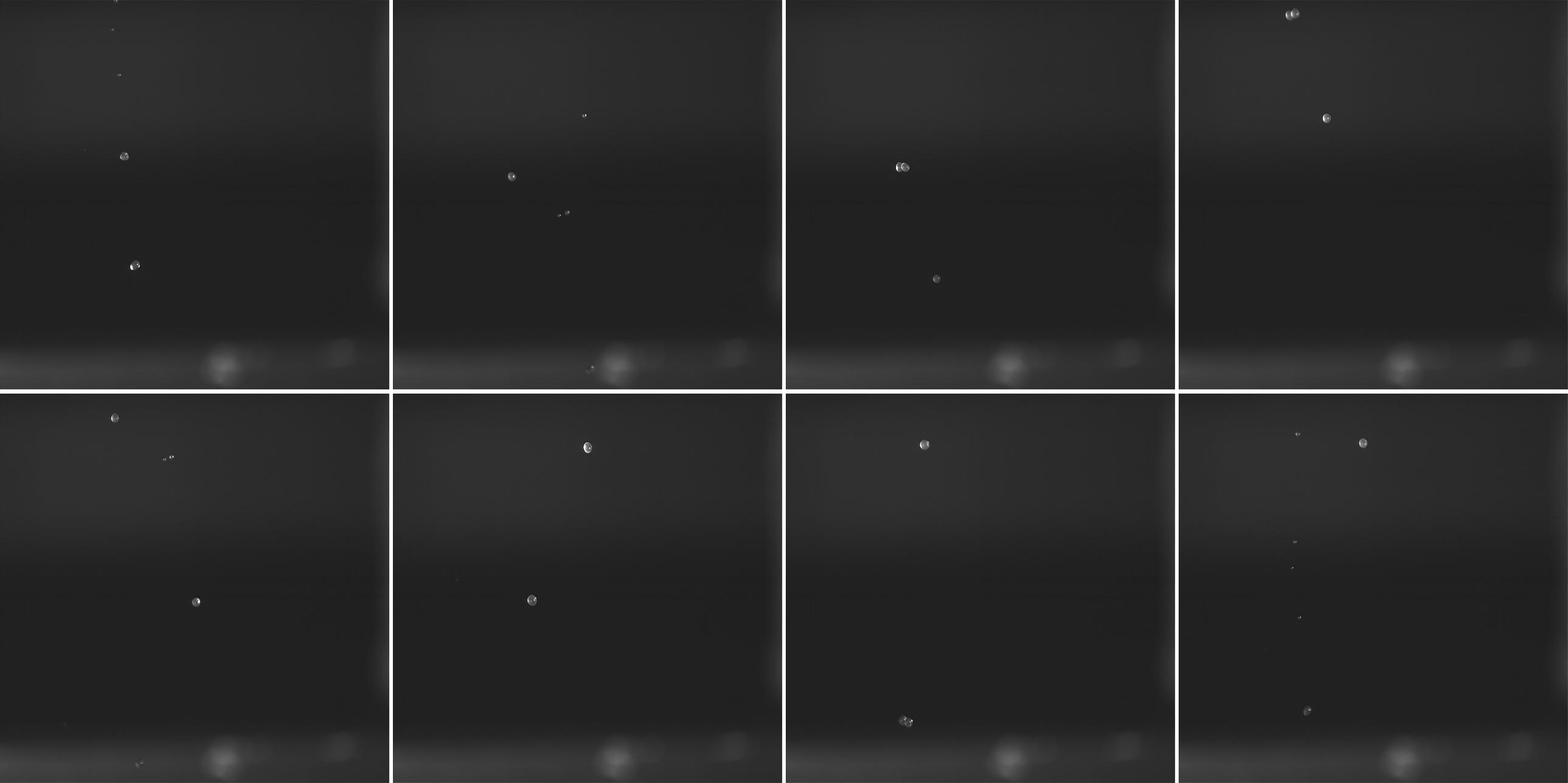}
\end{center}
\caption{Droplet images captured from a high-speed camera in the experiments.}
\label{fig.Dataset} 
\end{figure*}

The dataset comprises three sets: train (469 images), validation (125 images), and test (124 images). As depicted in Figure \ref{fig.Dataset}, the droplet size is substantially smaller than the overall image dimensions. For training the GAN droplet generator, we utilize 400 images from the training dataset as inputs. This GAN generator can subsequently produce up to 5000 synthetic droplet images, as discussed in Section \ref{sec.result}.

\section{Methods}
\label{sec.method}

Motivated by the need to address the bottleneck of data availability, we developed a model capable of generating droplet images through a generative AI framework. As illustrated in Fig. \ref{fig.Dataset}, the droplet sizes are notably small relative to the overall image dimensions. Therefore, our approach is generating high-resolution images from their low-resolution counterparts, utilizing the progressive mechanism \citep{karras2017progressive}. This section describes the mechanism and architecture of the model employed in the generation of droplet images.

\subsection{DropletGAN architecture}

\subsubsection{Generative mechanism}

\begin{figure*}[h]
\begin{center}
\includegraphics[width= 0.8\textwidth]{ 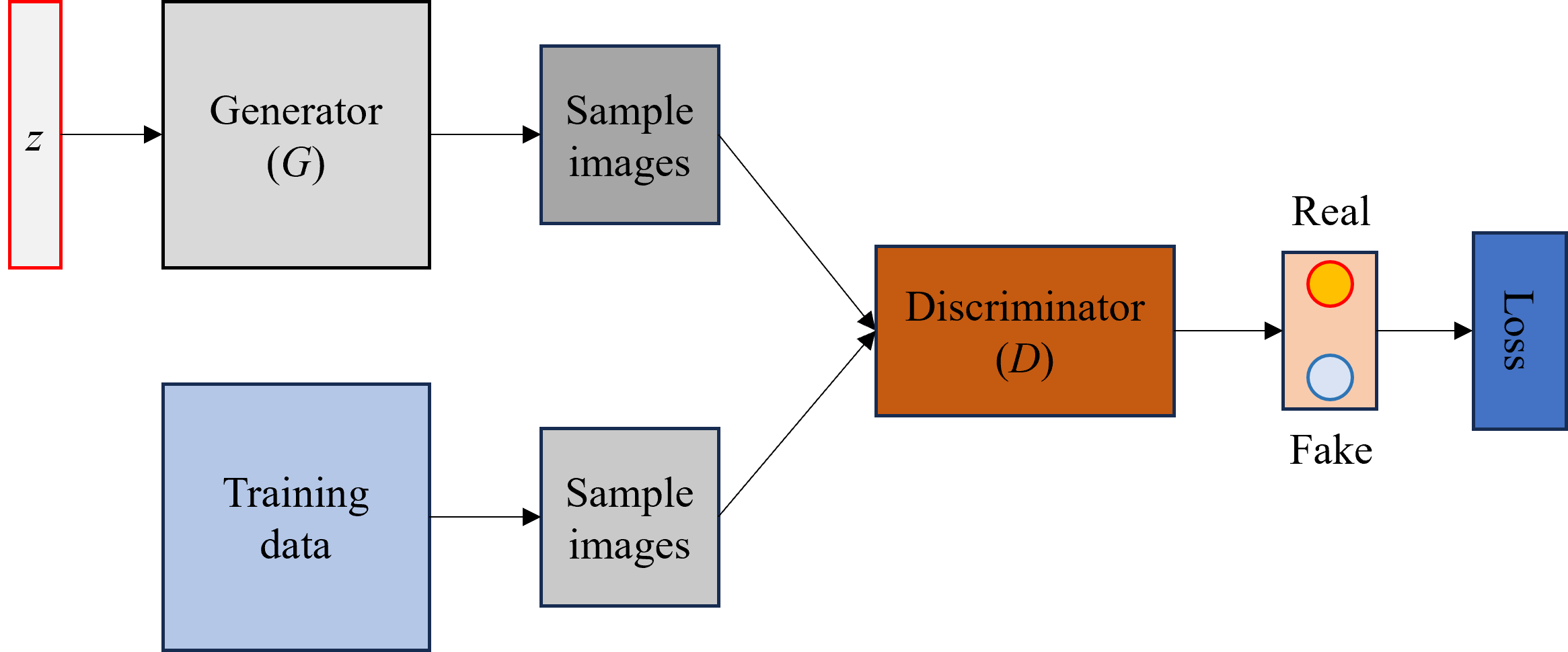}
\end{center}
\caption{The architecture of the GAN model. The generator and discriminator are trained together to finalize the best generator.}
\label{fig.GAN-model} 
\end{figure*}

The generative model consists of two networks: a generator $G$ and a discriminator $D$, as shown in Fig. \ref{fig.GAN-model}. In the training process, the generator tries to generate synthetic data from the noise vector $z$ that looks similar to real data to fool the discriminator. On the other hand, the discriminator tries to adjust its hyperparameters to distinguish the difference between fake and real data. The parameters of the generator and discriminator are updated constantly during the training. The training process can be formulated as a minimax game between the generator and discriminator as follows:
\begin{equation}
\label{loss}
\begin{split}
u_i=\min_{G} \max_{D} ([ &\mathbb{E}_{x \sim p_{\text{data}}(x)}[\log(D(x))]  + \\&\mathbb{E}_{z \sim p_z(z)}[\log(1 - D(G(z)))] )] .
\end{split}
\end{equation}
Here, $p_{\text{data}}(x)$ is the distribution of real data $x$, and \( p_z(z) \) is the distribution of the latent noise vector \( z \).

The model generates droplet images by following the progressive growth mechanism. Specifically, at the initial stages of training, the generator starts generating images with a low resolution, as illustrated in Fig. \ref{fig.Pro-GAN}. Then, new layers are incrementally added to the generator as the training progresses, allowing it to capture finer details in the images. Simultaneously, the discriminator also grows, learning to assess the increasing resolution of the generated images. In addition, it is worth noticing that all existing layers remain trainable during the growth process.

This incremental growth has several advantages. It leads to a more stable training process, mitigating issues such as mode collapse, where the generator produces a limited variety of samples. Additionally, it facilitates the generation of high-quality, high-resolution images, a significant improvement over traditional GANs. In the next sections, we will delve into the structure of the generator and discriminator networks.

\subsubsection{Generator}
\label{subsec.GAN}
The generator is a convolutional neural network (CNN) responsible for generating synthetic data. Given a random noise vector $z$ sampled from a latent space, the generator produces (fake) images $G(z)$ by using convolution layers that follow the distribution of real training images. Mathematically, this process can be expressed as $G:z \longrightarrow G(z; \theta_g)$, in which $\theta_g$ contains the parameters of the generator. 

\begin{figure*}[h]
\begin{center}
\includegraphics[width= 0.65\textwidth]{ 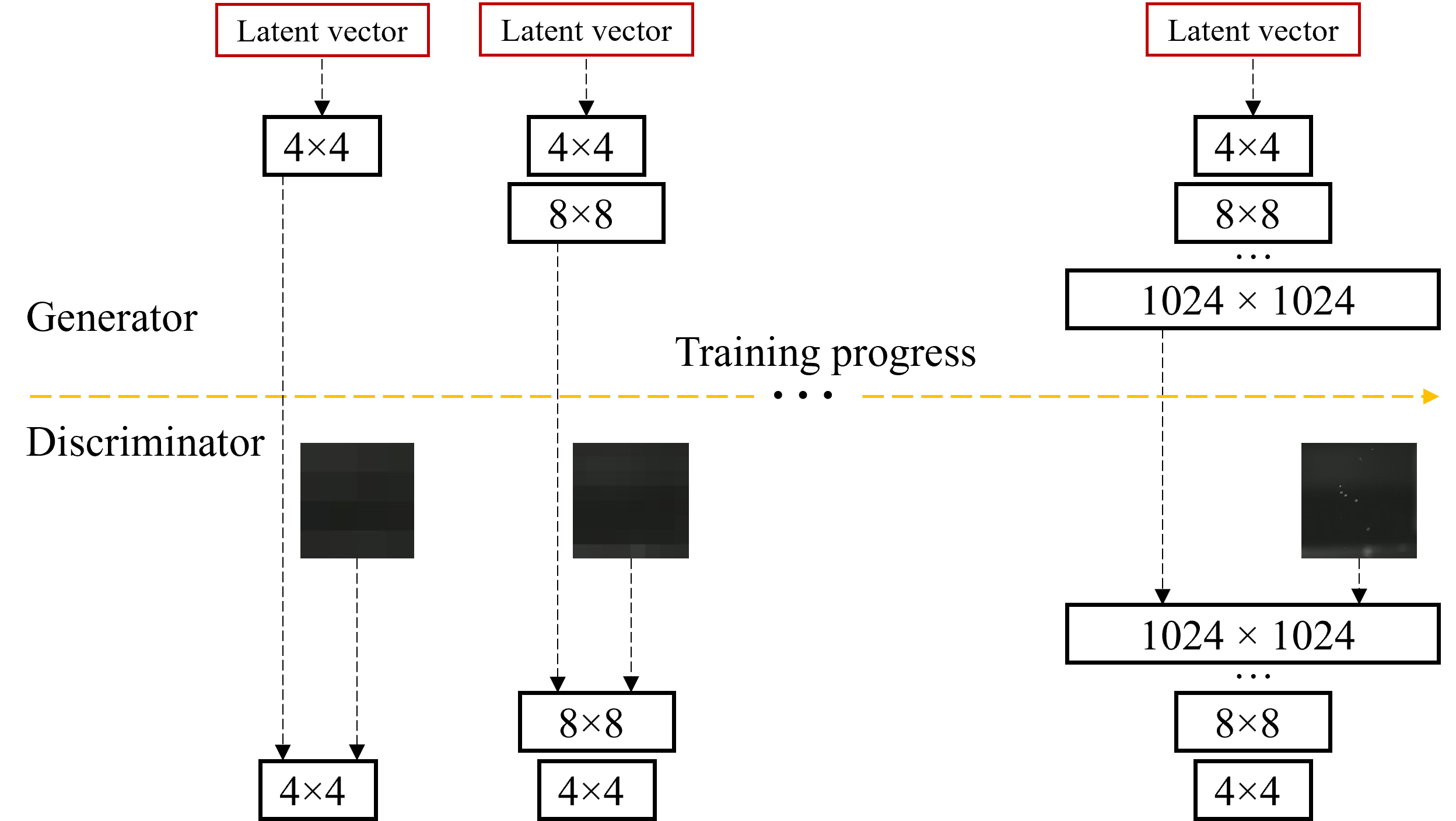}
\end{center}
\caption{The architecture of the generative model in the training process.}
\label{fig.Pro-GAN} 
\end{figure*} 

\begin{figure}[ht]
  \centering
  \begin{subfigure}{0.18\textwidth}
    \includegraphics[width=\linewidth]{ 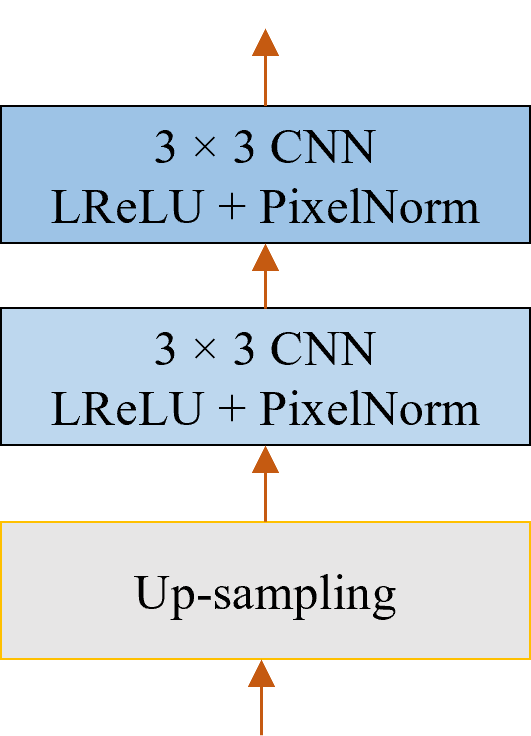}
    \caption{Generator block}
    \label{fig.generator_block}
  \end{subfigure}
  \hspace{1cm} 
  \begin{subfigure}{0.18\textwidth}
    \includegraphics[width=\linewidth]{ 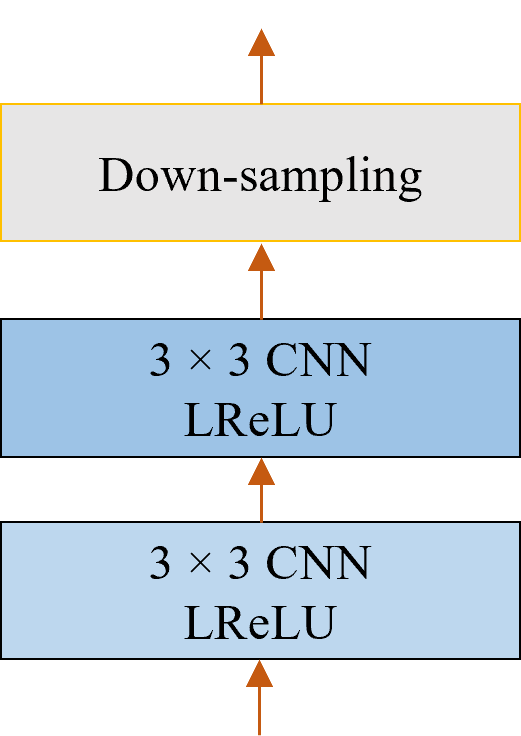}
    \caption{Discriminator block }
    \label{fig.discriminator_block}
  \end{subfigure}
  \caption{Generator and discriminator blocks in the progressive generative model.}
  \label{fig.generator_discriminator_block}
\end{figure}

Subsequently, the generator adds more layers to generate features with higher resolution in the next block as shown in Fig. \ref{fig.Pro-GAN}. The added layers in this process are illustrated in Fig. \ref{fig.generator_block}, which includes $3 \times 3$ convolution layers followed by LeakyReLU (LReLU) and PixelNorm activation functions. Let $g_i$ be a generic function that acts as the basic generator block, then the output of the block $g_i$ is 
\begin{equation}
A_i \in \mathbb{R}^{{2}^{i+1} \times {2}^{i+1} \times c_i}, 1\leq \forall{i} \leq  9,
\end{equation}
where $i \in N$, and $c_i$ is the number of channels in the generator block $i$-th. The progressive training process is repeated until the model generates features with the resolution of $1024 \times 1024$. In addition, at the end of the generator, a $1 \times 1$ convolution layer is applied to generate color images.

For instance, in the first block, the generator creates images with a resolution of $4 \times 4$ from the latent vector (noise) using convolution layers. Subsequently, the generator upscales the images generated in block 1 to a size of $8 \times 8$ in block 2. Similar to block 2, this image-generation process is reiterated until reaching the image size of $1024 \times 1024$. The progressive mechanism enables the generator to capture intricate features of droplets.

\subsubsection{Discriminator}
\label{subsec.Discriminator}
The discriminator is another convolutional neural network that evaluates whether a given data point is from the actual experiments or generated artificially by the generator. To do that, the discriminator also undergoes a progressive growth process. It starts with an architecture designed for evaluating low-resolution images and then expands its capacity as the resolution of generated images increases during training. 

As shown in Fig. \ref{fig.Pro-GAN}, given a generated data point $G(z)$, the discriminator outputs a probability $D(G(z))$ indicating the likelihood of $G(z)$ being a real sample. Mathematically, this can be expressed as $D:G(z) \longrightarrow D(G(z))$. The initial discriminator network is composed of a $3 \times 3$ convolution followed by a $4 \times 4$ convolution and a fully connected layer. Starting with a low-resolution image of size $4 \times 4$, the discriminator processes it using the initial networks and outputs the probability of the input image being real.

Similar to the generator's architecture, additional layers are added to the discriminator's network to handle larger images during the training. Let $d_i$ represent a basic discriminator block, the architecture of the block $d_i$ is described in Fig. \ref{fig.discriminator_block}. This discriminator block consists of two $3 \times 3$ convolution layers and a down-sampling layer.

During the training process, the discriminator is trained to minimize classification loss, whether the input is generated data or real data. The loss that measures the difference between the predicted probability of input being real and the actual label is formulated as follows:
\begin{equation*}
\label{Wasserstein-loss}
\begin{split}
L = &\mathbb{E}_{z \sim P_z}[D(G(z))] - \mathbb{E}_{x \sim P_{\text{data}}}[D(x)]  \quad \\ &+ \lambda \mathbb{E}_{\hat{x} \sim P_{\hat{x}}} \left[(\lVert \nabla_{\hat{x}} D(\hat{x}) \rVert_2 - 1)^2\right],
\end{split}
\end{equation*}
where \(P_{\text{data}}\) is the distribution of real samples, \(P_z\) is the distribution of generated samples, \(\hat{x}\) is a random sample along the straight lines between pairs of real and generated samples, and \(\lambda\) is a regularization parameter controlling the strength of the gradient penalty  \citep{gulrajani2017improved}.

When designing the discriminator architecture,  we employed a reverse engineering approach that aligns with the construction of the generator network. Beginning with an image, the input undergoes a series of neural network layers following the discriminator block until it reaches the real image size \citep{karras2017progressive}.

\subsection{Droplet Detection}
\label{sec.detection}

In this study, we introduce a droplet detection method leveraging the YOLOv8 object-detection model \citep{Jocher_YOLO_by_Ultralytics_2023}, which is designed for real-time object detection. Compared to other models, it stands out as an efficient and state-of-the-art model known for its high accuracy and fast processing speed. The model comprises two main components: a backbone and a head.

The \textbf{backbone} is an important part of the model, extracting meaningful features from the input image. It consists of 53 convolutional layers arranged in a modified CSPNet architecture \citep{redmon2018yolov3}. The design of the backbone network enables the model to capture intricate feature maps while remaining lightweight. 

Additionally, YOLOv8 integrates a Pyramid Pooling Fast module (SPPF), a feature that reduces computational complexity while preserving feature representation. It incorporates a Path Aggregation Network (PANet) that effectively aggregates features from different scales of the backbone. This allows the model to capture objects of various sizes and resolutions.

The detection \textbf{head} is responsible for predicting bounding boxes and class labels for objects in the input image. There are three sub-detection heads in the model, which enable the model to capture objects of different sizes. Furthermore, there is a decoupled head structure in each sub-detection head: one for predicting class labels and the other for predicting bounding boxes. This separation enables the model to focus independently on each task, thereby enhancing the accuracy of both bounding box predictions and object classification.

\section{Experiment and Evaluation Metrics}
\label{sec.metrics}

\subsection{Evaluation metrics for GAN}

To assess the quality of synthetic images produced by the generative model, we use the Fréchet inception distance (FID) \citep{heusel2017gans}. FID compares the distributions of features extracted from real images and those generated by our GAN. A lower FID score suggests the generated images are more realistic and diverse, closely resembling the distribution of real images. In FID, a pre-trained Inception V3 model is used to extract features from both real images (denoted by X) and generated images (denoted by G) \citep{szegedy2016rethinking}. This process essentially encodes the images into a lower-dimensional vector space capturing their characteristics. Then we model the data distribution for these features using a multivariate Gaussian distribution with mean $\mu$ and covariance $\Sigma$. The FID is computed using the following formula:

\begin{equation}
\label{FID}
\text{FID(X,G)} = ||\mu_X - \mu_G||^2 + \text{Tr}(\Sigma_X + \Sigma_G - 2(\Sigma_X^{\frac{1}{2}} \Sigma_G^{\frac{1}{2}})).
\end{equation}

Here, $||\mu_X - \mu_G||^2$ represents the squared Euclidean distance between the mean vectors ($\mu_X$ and $\mu_G$) of the real and generated features. $\text{Tr}(\Sigma_X + \Sigma_G - 2(\Sigma_X^{\frac{1}{2}} \Sigma_G^{\frac{1}{2}}))$ represents the trace of the sum of the covariance matrices ($\Sigma_X$ and $\Sigma_G$) of the real and generated data, minus twice the product of their square roots. $\Sigma_X^{\frac{1}{2}}$ denotes the matrix square root of the covariance matrix for the real data. $\Sigma_G^{\frac{1}{2}}$ denotes the matrix square root of the covariance matrix for the generated data.

In addition, the fidelity and diversity of the synthesized images generated by the generator are evaluated by precision and recall metrics through droplet detection. High precision indicates that the generated images closely resemble real images on average, while high recall suggests that the generator can produce any sample present in the training dataset \citep{lucic2018gans, xu2018empirical}.

\subsection{Evaluation metrics for Object detection}
In the evaluation process, the trained droplet detector takes in images, detects droplets in the images, and then generates bounding boxes around detected droplets. To evaluate the performance of the detector, we subsequently compared the predictions with the corresponding ground truths. This involves a detailed analysis of the spatial overlap between predicted and ground-truth bounding boxes.

To facilitate the evaluation process, we defined specific parameters as follows:
\begin{itemize}
    \item True Positive (TP) is the number of correctly predicted droplet bounding boxes compared with the ground truth bounding boxes of droplets. 
    \item False Positive (FP) is the number of incorrectly predicted droplet bounding boxes compared with the ground truth bounding boxes of droplets. 
    \item False Negative (FN) is the number of predicted droplet bounding boxes when the model fails to detect actual droplets present in the image.
\end{itemize}

Based on the definition above, the evaluation metrics are defined as 
\begin{equation}
    \text{Precision} = \frac{\text{TP}}{\text{TP + FP}}
\end{equation}

\begin{equation}
    \text{Recall} = \frac{\text{TP}}{\text{TP + FN}}
\end{equation}
where Precision represents the accuracy of positive predictions that is computed as the ratio of TP to the sum of TP and FP. Recall, on the other hand, is the ratio of TP to the sum of TP and FN, indicating the model's ability to capture all positive instances.

Beyond using Precision and Recall metrics, we also use Average Precision (AP) to evaluate the performance of the model. Average Precision is a comprehensive metric that considers precision-recall trade-offs at various confidence thresholds. A higher AP signifies a more robust and reliable droplet detection model. It involves calculating the area under the precision-recall curve and is defined as follows:

\begin{equation}[h]
    \text{AP} = \int_{0}^{1} \text{Precision}(\text{Recall}) \, d(\text{Recall})
\end{equation}

\section{Results and discussion}
\label{sec.result}

\begin{figure*}[h]
\begin{center}
\includegraphics[width= 0.85\textwidth]{ 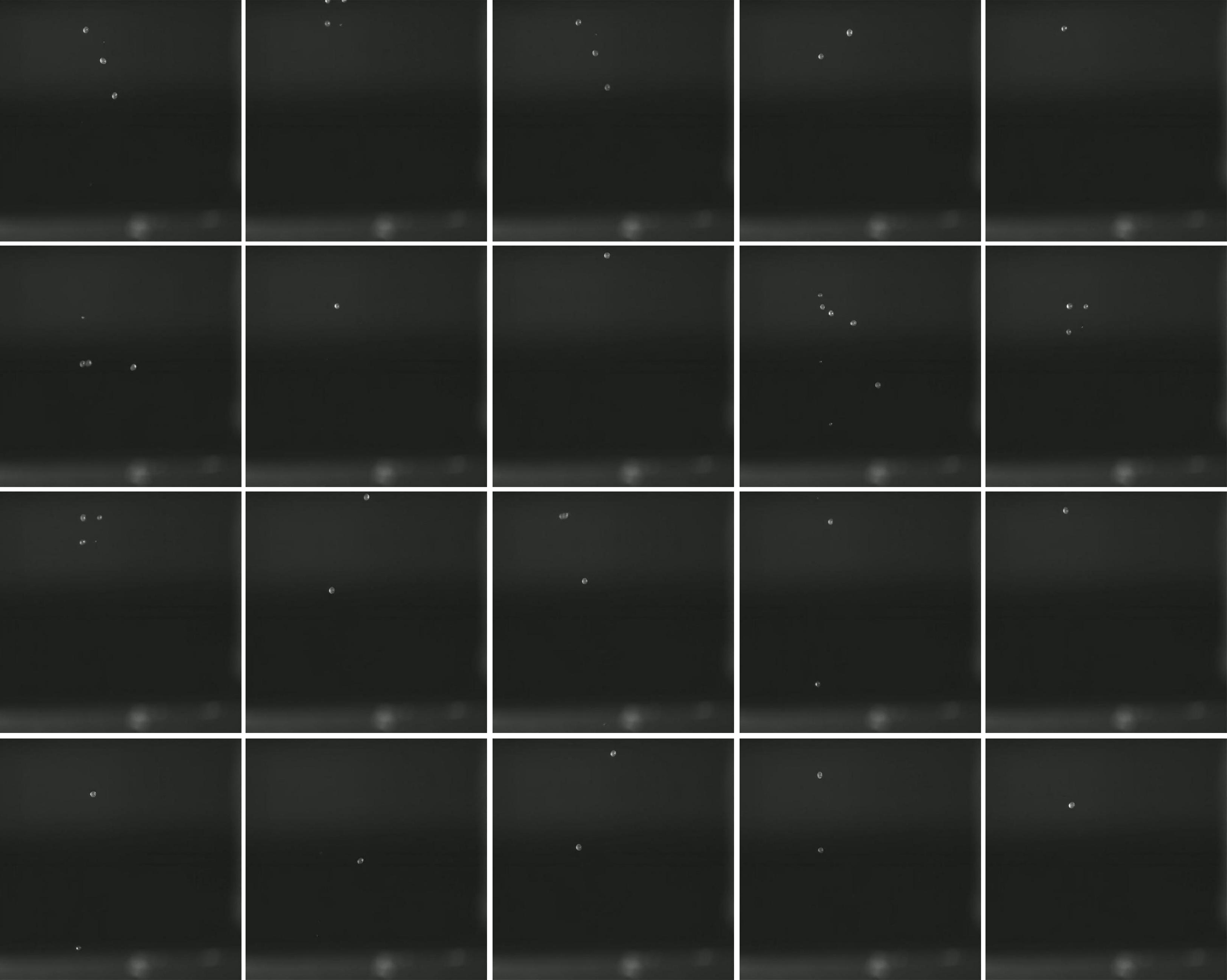}
\end{center}
\caption{Droplet images produced by the generative model.}
\label{fig.Generated-images} 
\end{figure*} 

\subsection{Synthetic Dataset with the generative model}

After training with real images, the generator adeptly generated more than 5000 synthetic images with a resolution of $1024 \times 1024$. This volume of droplet images would require tremendous resources if generated via experiments. However, not all of these images have high-quality standards, hence, we selected around 1500 images for our subsequent experiments. The representations of these images are illustrated in Fig. \ref{fig.Generated-images}.

The calculated FID score is 11.29, which is small enough to generate good-quality images \citep{zhang2019self, brock2018large}. From observation, the generated images exhibit a notable resemblance to real ones, illustrating not only in terms of resolution but also in capturing intricate details, including small droplets. The generative model demonstrates its capability to produce realistic droplet images, thereby enriching the database with a diverse array of visual features.

\begin{figure}[h]
\centering
  \begin{subfigure}{0.4\textwidth}
    \includegraphics[width=\linewidth]{ 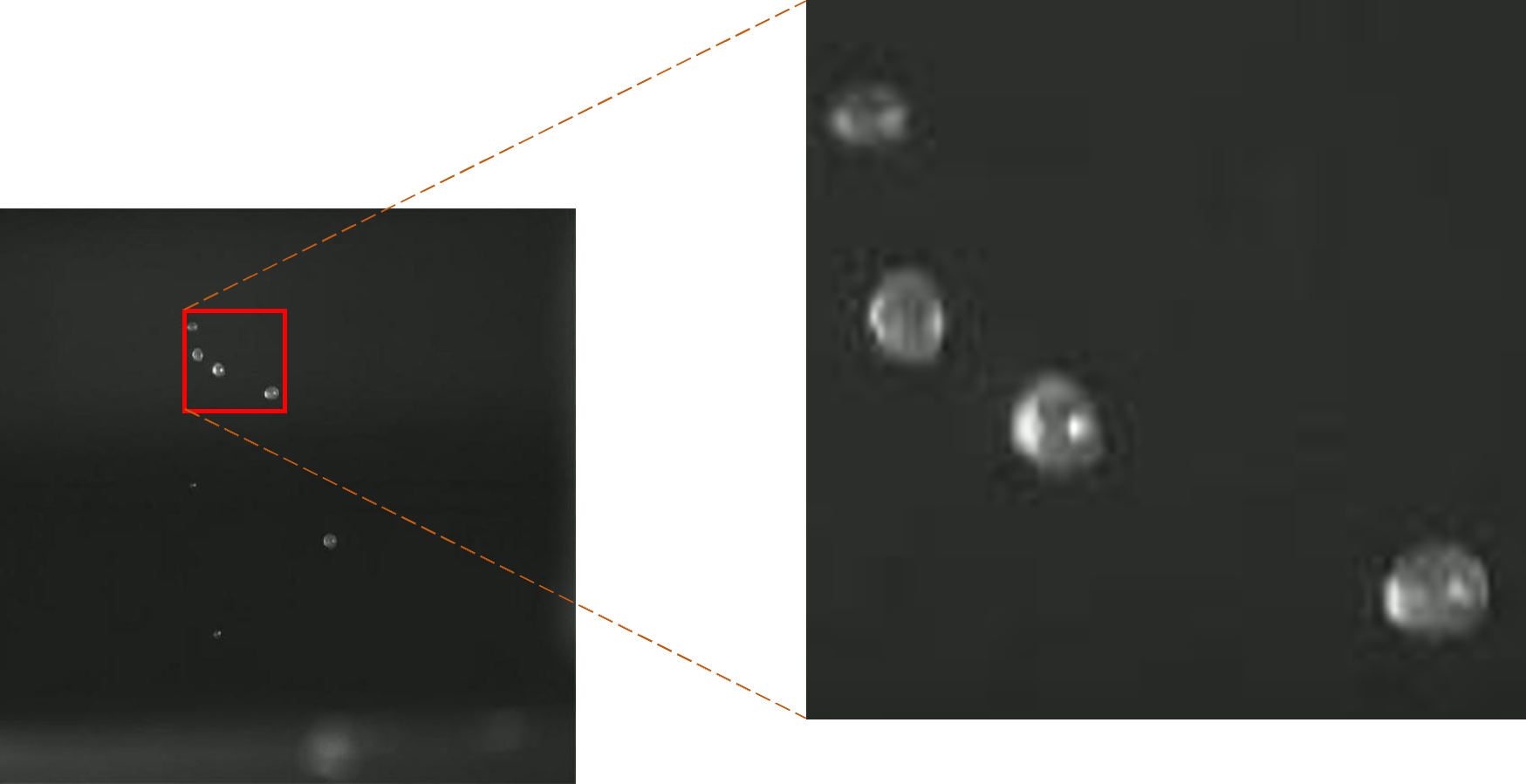}
    \caption{Fake comparison}
    \label{fig.Fake-comparison}
  \end{subfigure}
  \hfill
  \begin{subfigure}{0.4\textwidth}
    \includegraphics[width=\linewidth]{ 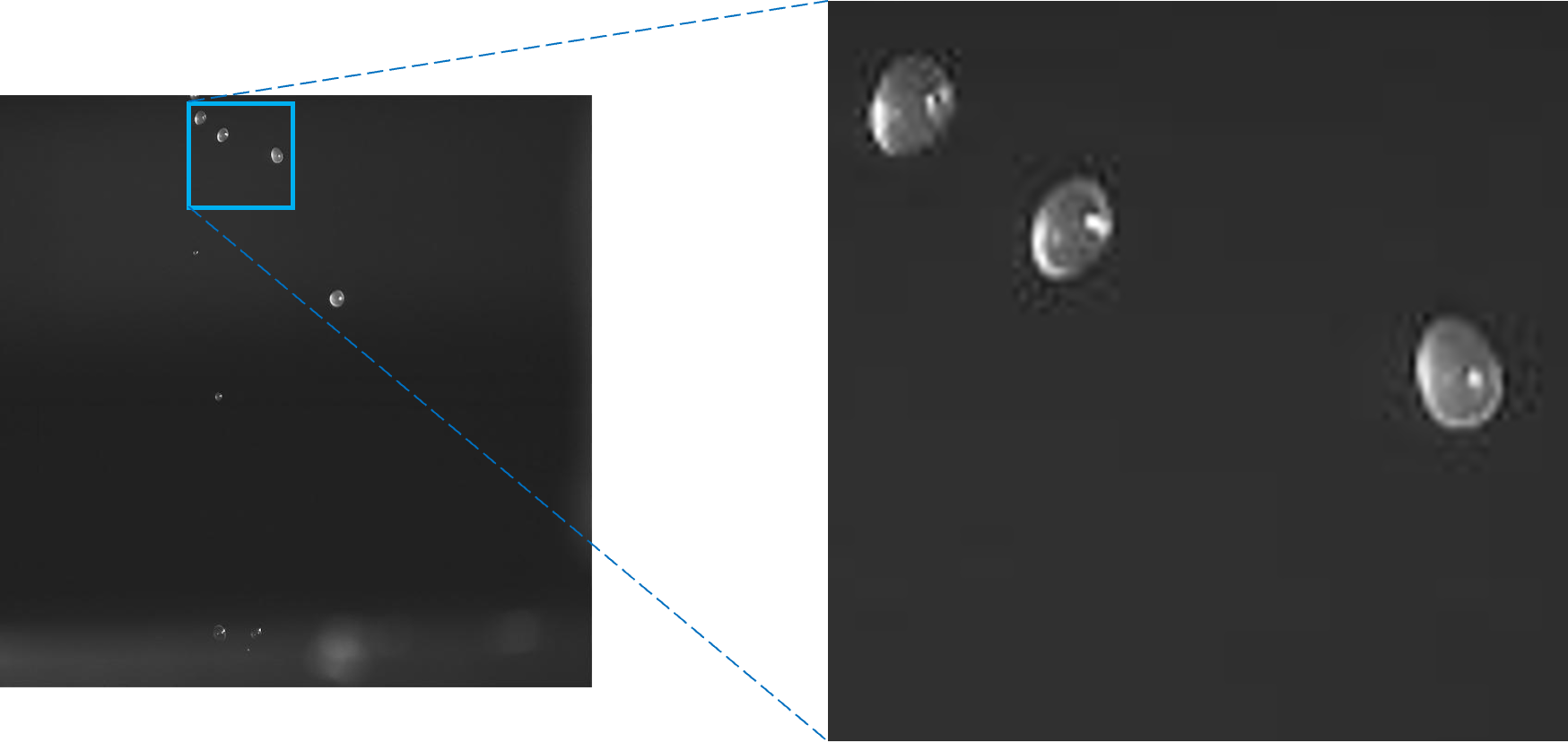}
    \caption{Real comparison}
    \label{fig.Real-comparison}
  \end{subfigure}
  \caption{Comparison of generated images and real images.}
  \label{fig.Real-fake-comparison}
\end{figure}

For a more intuitive comparison between generated and real images, we cropped the section containing droplets in the images and presented it in Fig. \ref{fig.Real-fake-comparison}. Specifically, Figure \ref{fig.Fake-comparison} showcases a synthetic image produced by the generative AI model with its corresponding zoomed-in area, while Figure \ref{fig.Real-comparison} displays a real image from the experiment accompanied by its zoom area. As observed, the zoomed areas reveal subtle distinctions between images, in which the real image shows that the contrast of droplets in the image is higher than in the synthetic image.

\begin{table*}[ht]
\centering
\caption{Evaluate the quality of generated images by using Precision (Pre.) and Recall (Rec.) metrics. We used the original dataset (469 training samples) and the synthetic dataset (500 samples) for the comparison.}
\label{tab.real_fake_quality}
\begin{tabular}{l|cccc|cccc}
\toprule
\multirow{2}{*}{\textbf{Data}} & \multicolumn{4}{c|}{\textbf{Validation}} & \multicolumn{4}{c}{\textbf{Test}} \\
\cmidrule(lr){2-5} \cmidrule(lr){6-9}
 & Pre. & Rec. & mAP50 & mAP50-95 & Pre. & Rec. & mAP50 & mAP50-95 \\
\midrule \addlinespace[-0.25ex] \midrule
Real (469)& 0.920 & 0.898 & 0.954 & 0.635 & 0.891 & 0.877 & 0.950 & 0.636 \\
Synthetic (500) & 0.909 & 0.888 & 0.949 & 0.623 & 0.87 & 0.874 & 0.941 & 0.612 \\
\bottomrule
\end{tabular}
\end{table*}

In addition to the visual results, we conducted a quantitative analysis using metrics discussed in Section \ref{sec.metrics}. Table \ref{tab.real_fake_quality} details the training of YOLOv8n with 469 real images and 500 synthetic images separately. Moreover, it presents the results of testing these models on validation and test sets. As shown in Table \ref{tab.real_fake_quality}, Precision and Recall scores are slightly lower when using synthetic images compared to real images, though the difference is not significant. Specifically, the model trained with real data achieves Precision and Recall scores of 0.891 and 0.877, respectively, on the test set, while the model trained with synthetic data achieves a Precision score of 0.87 and a Recall score of 0.874.

\subsection{Droplet detection results}

After obtaining the synthetic dataset, a series of experiments were conducted to improve droplet detection performance. The results are listed in Table \ref{tab.detection_results}. Initially, YOLOv8n was trained using the real dataset, comprising 469 images. As shown in Table \ref{tab.detection_results}, the model achieved an mAP50 of 0.950 and an mAP50-90 of 0.636 on the test set. On the validation set, the corresponding values were 0.954 for mAP50 and 0.635 for mAP50-95.

To leverage the synthetic dataset, we utilized the pre-trained model mentioned above to label the data. It's important to note that the labeled data at this stage may not be perfectly accurate, therefore, we labeled the dataset based on the initial labeled data.

\begin{table*}[h]
\centering
\caption{Comparison of droplet detection results between the original dataset and augmented dataset. In this table, ``Pre." and ``Rec." represent Precision and Recall metrics, respectively. Additionally, the detection performance on the real dataset is denoted by (R).}
\label{tab.detection_results}
\begin{tabular}{l|cccc|cccc}
\toprule
\multirow{2}{*}{\textbf{Data}} & \multicolumn{4}{c|}{\textbf{Validation}} & \multicolumn{4}{c}{\textbf{Test}} \\
\cmidrule(lr){2-5} \cmidrule(lr){6-9}
 & Pre. & Rec. & mAP50 & mAP50-95 & Pre. & Rec. & mAP50 & mAP50-95 \\
\midrule \addlinespace[-0.25ex] \midrule
469 (R)& 0.920 & 0.898 & 0.954 & 0.635 & 0.891 & 0.877 & 0.950 & 0.636 \\
969 & 0.955 & 0.89 & 0.96 & 0.685 & 0.956 & 0.888 & 0.970 & 0.683 \\
1469 & 0.964 & 0.916 & 0.974 & 0.698 & 0.948 & 0.929 & 0.977 & 0.704 \\
1783 & 0.938 & 0.932 & 0.966 & 0.717 & 0.938 & 0.927 & 0.973 & 0.703 \\
2504 & 0.969 & 0.962 & 0.988 & 0.737 & 0.951 & 0.970 & 0.988 & 0.721 \\
\bottomrule
\end{tabular}
\end{table*}

To assess the droplet detection effectiveness, we gradually increased the number of training samples by adding synthetic (fake) images, resulting in sizes of 969, 1469, 1783, and 2504 images. Importantly, we kept the number of samples consistent in the validation and test sets.

As observed, the detector's performance improves with the addition of synthetic samples. For example, when combining 469 real images with 500 synthetic images, following the comparison in Table \ref{tab.real_fake_quality}, the mAP50-95 scores increased by 7.38\% on the test set and 7.87\% on the validation set. Furthermore, in the case of a 2504-sample training, the mAP50 increased from 0.950 to 0.988, and mAP50-95 increased from 0.636 to 0.721 on the test set compared to the real-sample training. Similarly, a consistent trend is observed in the validation set, where mAP50 increased to 0.988, and mAP50-95 increased to 0.737.

\begin{figure}[H]
  \centering
  \begin{subfigure}{0.4\textwidth}
    \includegraphics[width=\linewidth]{ 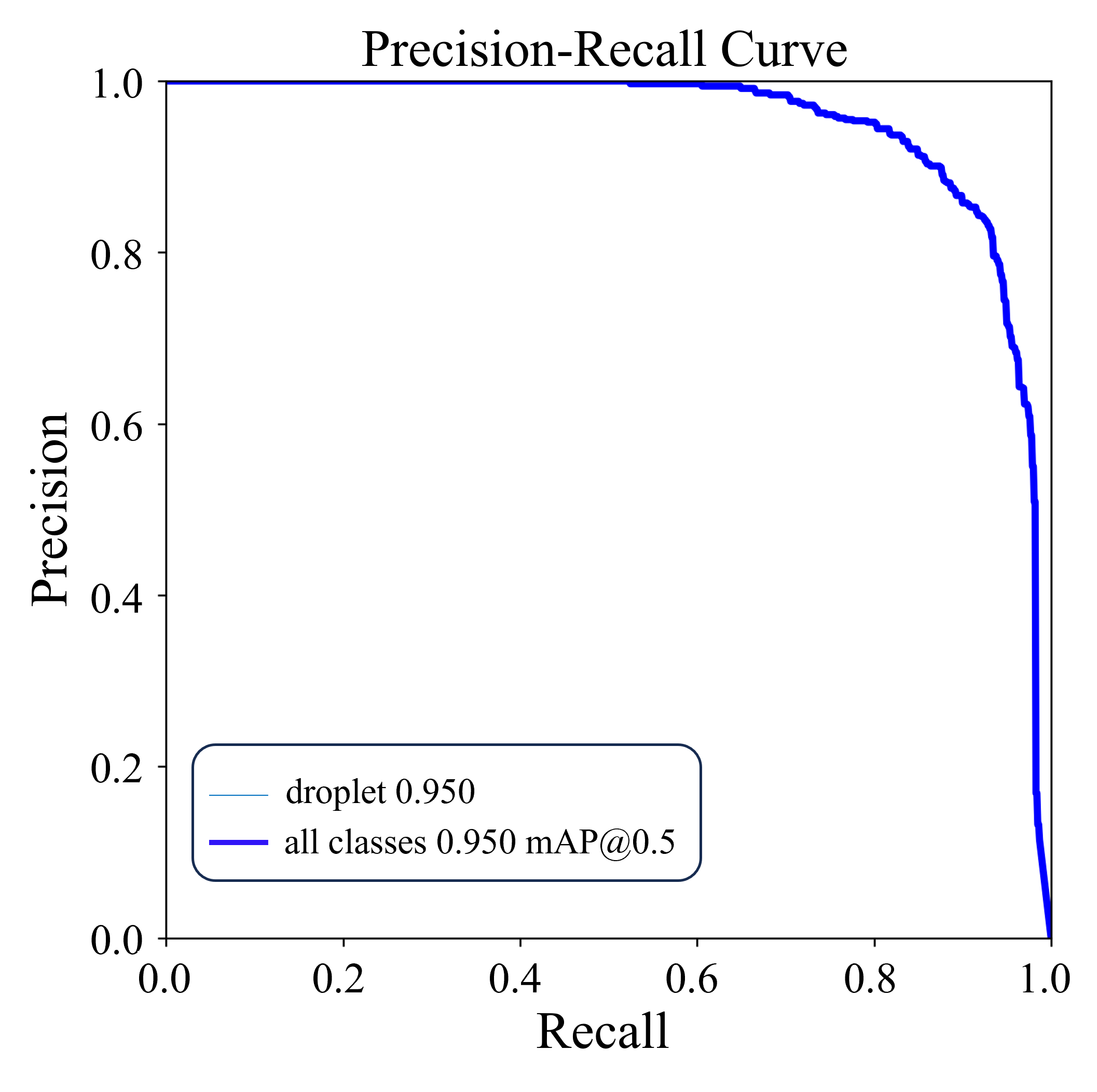}
    \caption{Precision-Recall curve on original dataset (using only real dataset)}
    \label{fig.PR-curve496}
  \end{subfigure}
  \hfill
  \begin{subfigure}{0.4\textwidth}
    \includegraphics[width=\linewidth]{ 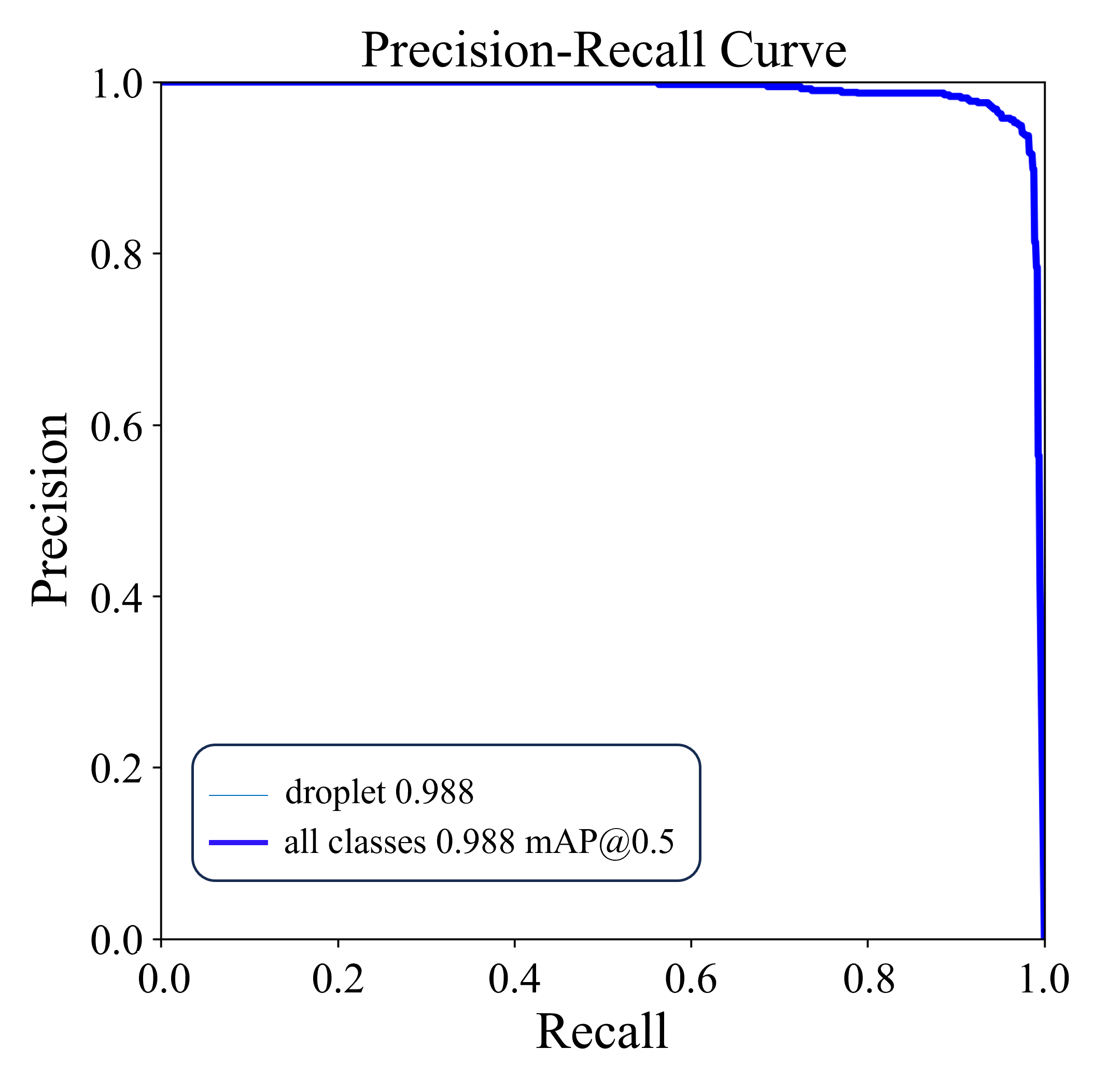}
    \caption{Precision-Recall curve on augmented dataset (using augmented dataset)}
    \label{fig.PR-curve2504}
  \end{subfigure}
  \caption{Comparison of droplet detection performance on the test set in terms of the Precision-Recall curve when utilizing an augmented dataset.}
  \label{fig.PR-curve}
\end{figure}

In addition to the results presented in Table \ref{tab.detection_results}, we compare Precision-Recall (PR) curves between the original training (only real images) and the training with 2504 samples (real images + synthetic images), as shown in Fig. \ref{fig.PR-curve}. Notably, the area under the PR curve for the original training is smaller compared to that of the augmented training. Consequently, incorporating generated data into the training set yields improved results. 

\begin{figure*}[h]
\begin{center}
\includegraphics[width= 0.85\textwidth]{ 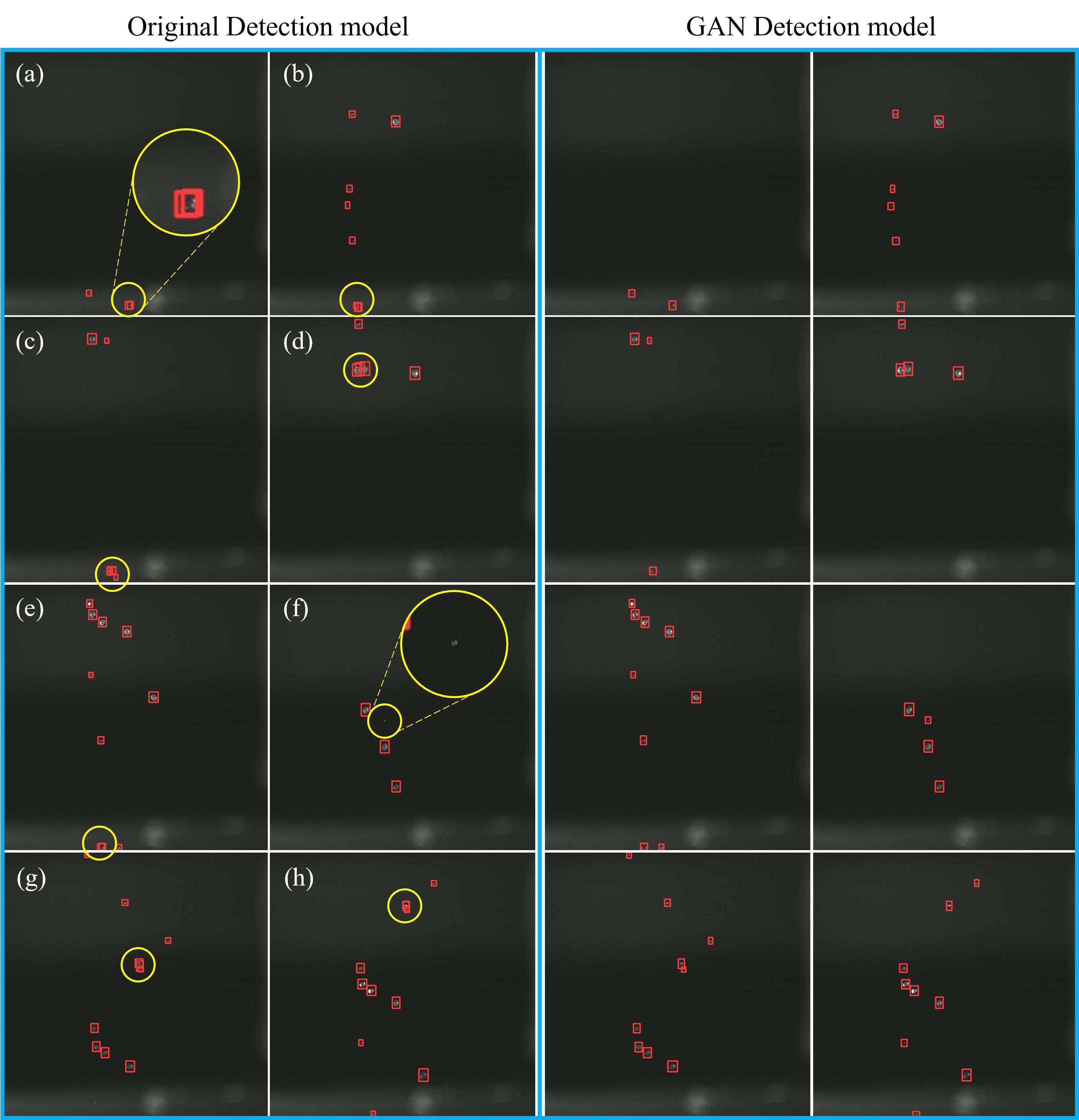}
\end{center}
\caption{Prediction comparison based on the original training data (469 samples) and augmented training data (2504 samples). The initial two columns display predictions from the original training, while the last two columns show predictions from the augmented data, corresponding to their counterparts in the first two columns. Typically, the inaccurate predictions are highlighted by yellow circles. }
\label{fig.Prediction} 
\end{figure*}

Fig.~\ref{fig.Prediction} illustrates the comparative efficacy of droplet detection of the two models, one was trained with only real droplet images and the other was trained with mixed real and synthetic droplets. The first two columns depict the droplet detection results produced by the model trained with only real droplet images (as indicated by the first train case in Table \ref{tab.detection_results}). The last two columns show the droplet detection results on the same images produced by the model trained with the 2504-sample dataset (mixed real and synthetic images). Moreover, the yellow circles highlight inaccuracies in the detection by the first model.

In particular, when utilizing the augmentation dataset (mixed real and synthetic images) for training, the GAN model detects droplets more precisely as compared to the original model trained with just real droplet images. As illustrated in the zoomed region of Fig.~\ref{fig.Prediction}a, the original model predicted one droplet with two bounding boxes, while the GAN detection model predicted exactly one bounding box for a single droplet. This comparison can be observed similarly in Figs.~\ref{fig.Prediction}b, c, d, e, g, and h. In addition, the GAN detection model can detect all droplets in the images, including small ones that the original model misses, as demonstrated in Fig.~\ref{fig.Prediction}f.

\section{Conclusion}
\label{conclusion}
This work introduces a novel approach to streamline spray system optimization through a data augmentation method. We proposed DropletGAN to create a synthetic dataset. The DropletGAN model consists of a generator and a discriminator: the generator produces images from noise to fool the discriminator; the discriminator tries to distinguish images generated by the generator and the real images. 

To train the generative AI model, we collect data from experiments using a high-speed camera to capture droplet images. Instead of initially training the generator to produce high-resolution images, the training process begins with the generator creating images at a small resolution of $4 \times 4$ and gradually increasing it up to $1024 \times 1024$. This progressive training facilitates the model in learning features at different scales, enhancing its stability and ability to generate high-quality images. As a result, the Fréchet inception distance score of the generated dataset is 11.29. This score illustrates that these images are good-quality images compared to real images.

After obtaining the synthesized dataset, we labeled it using a pre-trained model. To accomplish this, we trained YOLOv8-n with the original images and then used the pre-trained model to label the synthetic dataset. Notably, the labeled synthetic data from this process is not of high quality. Therefore, we re-labeled the dataset based on the automatically labeled data.

After getting the dataset, we conducted a series of experiments for droplet detection with models trained with different datasets, including one with only real images and several with a mixture of real and synthetic images. First, we trained the droplet detection model with the original dataset composed of only real images. After that, we gradually added an amount of synthetic data to the initial training dataset. The results show that the detection performance is improved as the number of synthetic images increases. Specifically, the model achieved mAP50 and mAP50-90 scores of 0.950 and 0.636 on the original dataset while these metrics in the case of mixed-data training are 0.988 and 0.721, respectively.

Although the model can generate more datasets, they are limited to a few varieties. Therefore, in the future, we aim to find a way to generate a dataset with greater diversity. We plan to build a larger database for droplet images based on the generative model and use our model to automate the labeling process. Additionally, we will apply the droplet detector to estimate droplet characteristics during the nozzle optimization design process.

\bibliographystyle{IEEEtran}  
\bibliography{Cite}

\end{document}